\documentclass[runningheads]{llncs}
\usepackage{graphicx}
\usepackage{tikz}
\usepackage{comment}
\usepackage{amsmath,amssymb} % define this before the line numbering.
\usepackage{color}
\usepackage{multirow}

\usepackage[accsupp]{axessibility}  % Improves PDF readability for those with disabilities.

\begin{document}
\pagestyle{headings}
\mainmatter
\def\ECCVSubNumber{100}  % Insert your submission number here

\title{Two-Aspect Information Fusion Model For ABAW4 Multi-task Challenge} % Replace with your title

% INITIAL SUBMISSION 
%\begin{comment}
\titlerunning{Two-Aspect Information Fusion Model} 
\authorrunning{H. Sun, et al} 
\author{Haiyang Sun$^{1,2}$, Zheng Lian$^{1}$, Bin Liu$^{1^\dag}$, Jianhua Tao$^{1,2,3^\dag}$, Licai Sun$^{1,2}$, Cong Cai$^{1,2}$}
\institute{
	$^1$National Laboratory of Pattern Recognition, Institute of Automation, Chinese Academy of Sciences\\
	$^2$School of Artificial Intelligence, University of Chinese Academy of Sciences\\
	$^3$CAS Center for Excellence in Brain Science and Intelligence Technology
	\email{\{sunhaiyang2021, lianzheng2016\}@ia.ac.cn \\ \{liubin, jhtao\}@nlpr.ia.ac.cn}}
%\end{comment}
%******************

% CAMERA READY SUBMISSION
\begin{comment}
\titlerunning{Abbreviated paper title}
% If the paper title is too long for the running head, you can set
% an abbreviated paper title here
%
\author{First Author\inst{1}\orcidID{0000-1111-2222-3333} \and
Second Author\inst{2,3}\orcidID{1111-2222-3333-4444} \and
Third Author\inst{3}\orcidID{2222--3333-4444-5555}}
%
\authorrunning{F. Author et al.}
% First names are abbreviated in the running head.
% If there are more than two authors, 'et al.' is used.
%
\institute{Princeton University, Princeton NJ 08544, USA \and
Springer Heidelberg, Tiergartenstr. 17, 69121 Heidelberg, Germany
\email{lncs@springer.com}\\
\url{http://www.springer.com/gp/computer-science/lncs} \and
ABC Institute, Rupert-Karls-University Heidelberg, Heidelberg, Germany\\
\email{\{abc,lncs\}@uni-heidelberg.de}}
\end{comment}
%******************
\maketitle

\begin{abstract}
In this paper, we propose the solution to the Multi-Task Learning (MTL) Challenge of the 4th Affective Behavior Analysis in-the-wild (ABAW) competition. The task of ABAW is to predict frame-level emotion descriptors from videos: discrete emotional state; valence and arousal; and action units. Although researchers have proposed several approaches and achieved promising results in ABAW, current works in this task rarely consider interactions between different emotion descriptors. To this end, we propose a novel end to end architecture to achieve full integration of different types of information. Experimental results demonstrate the effectiveness of our proposed solution.
\keywords{ABAW4 Multi-task Challenge, information fusion, emotion recognition, action unit detection}
\end{abstract}

\section{Introduction}
Automatic emotion recognition techniques have been an important task in affective computing. Previously, Ekman et al. \cite{ekman1969repertoire} divided facial emotion descriptors into two categories: sign vehicles and messages. The sign vehicles are determined by facial movements, which are better fit with action units (AU) detection. The messages are more concerned with the person observing the face or getting the message, which are better fit with facial expressions (EXPR), valence and arousal (VA) prediction.

Deng \cite{DBLP:journals/corr/abs-2203-12845} assumed that estimates of EXPR and VA vary across different observers, but AU is easier to reach consensus. This phenomenon indicates that the sign vehicles and messages express two aspects of emotion. Therefore, the author use different modules to predict them separately. However, observing a smiling but frowning face, the intensity of frowning and smiling influences the change of action units to some extent; while the amplitude of the eyebrow and lip movements influences whether the expressions are positive, and the intensity of emotional states. Therefore, we assume that the results of sign vehicles and messages are are mutually influenced.

In this paper, we present our solution to the ABAW competition. Specifically, we leverage the ROI Feature Extraction module in \cite{DBLP:journals/corr/abs-2203-12845,jacob2021facial} to capture the facial emotion information. Afterwards, we leverage interactions between sign vehicles and messages to achieve better performance in the ABAW competition.Our main contributions are as follows:
\begin{enumerate}
	\item We make the model compute two aspects of the information by a two-way computation, to represent the information of sign vehicles and messages.
	\item We increase the information interoperability between these two aspects to better integrate them into multiple tasks.
	\item Our method showed superior performance than baseline model.
\end{enumerate}

\section{Methodology}
\begin{figure}
	\centering
	\includegraphics[width=1\linewidth]{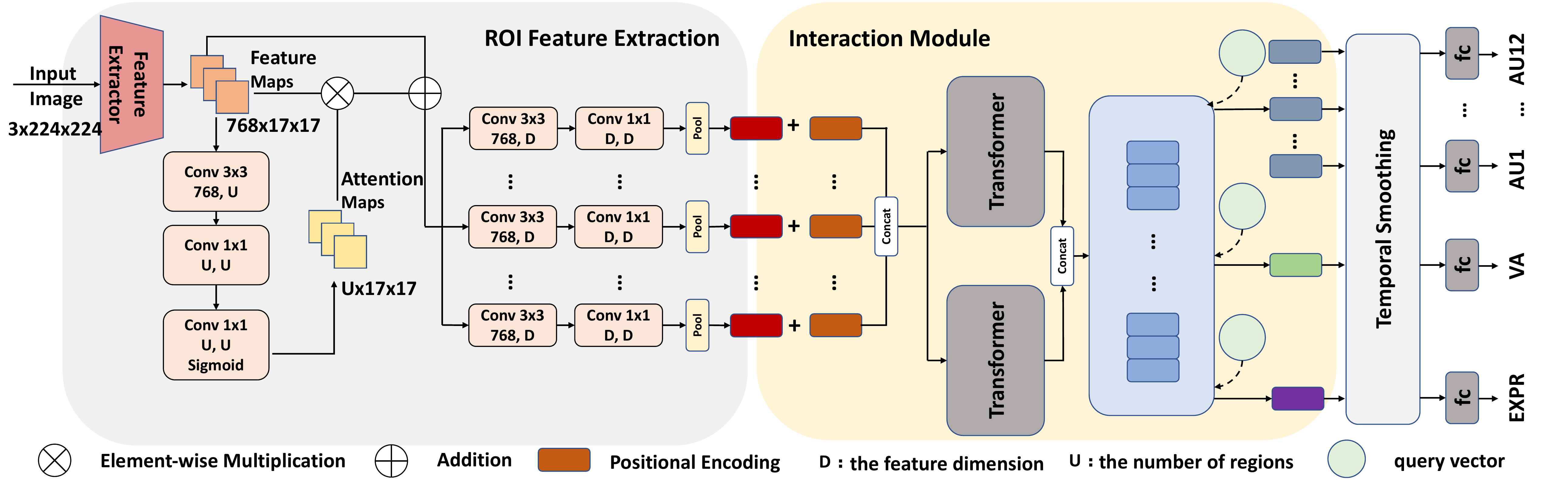}
	\caption{ROI Feature Extraction and Interaction Module.}
	\label{fig:model_architecture1}
\end{figure}

Our model has three main components: ROI Feature Extraction, Interaction Module, and Temporal Smoothing Module. The overall framework is shown in Figure \ref{fig:model_architecture1}.

\subsection{ROI Feature Extraction}
The main function of ROI Feature Extraction is to encode important information in the images. We adopt the same module in \cite{DBLP:journals/corr/abs-2203-12845}. The feature extractor is an Inception V3 model, images are fed into it and generate feature maps. After that, feature maps generate spatial attention maps for the model's regions of interest by using three convolutional layers and a sigmoid calculation. The attention maps are fused with feature maps to generate feature vectors of different regions by different encoding modules shown by red tensors. More details about ROI Feature Extraction are in \cite{jacob2021facial}.

\subsection{Interaction Module}
The main function of Interaction Module is to extract information from different semantic spaces, then provide a more comprehensive representation of emotion descriptors for different tasks through information interaction operations. 
Coding through the ROI Feature Extraction module, each image is transformed into $U$ feature vectors to represent the information of the entire face. These representations are combined with positional encoding vectors and fed into two Transformer blocks to learn information from two perspectives. The outputs of these two blocks are concatenated together as an overall information representation shown by blue tensors. We assign a separate trainable query vector to each task. Each task separately performs query operation on the overall representation and integrates information from different perspectives.

\subsection{Temporal Smoothing and Classifier Module}
The main function of Temporal Smoothing Module is to add the time domain information to the feature vector. Since the previous modules are calculated on a single image, the temporal information of the video data cannot be taken into account. Therefore, we perform a temporal smoothing operation on the encoded features. After extracting feature vectors from a piece of video data, for the feature vector $v_t$ at the time step $t$, we smoothed it with this function:
\begin{align}
f_t = \frac{1}{1+\mu}(v_t+\mu f_{t-1})
\end{align}
where $f_t$ is the feature that is fed to the classifier at time $t$. Unlike \cite{DBLP:journals/corr/abs-2203-12845}, we add the temporal smoothing operation in the training phase and assign a trainable $\mu$ to each task. It is important to emphasize that we set a trainable $f_{-1}$ for each task for the case $t=0$. To better train $\mu$, we discard videos with less than 10 frames of data.
Finally, we feed the smoothed features into simple FC layers for classification or prediction.

\subsection{Losses}
We use BCEloss as the inference loss for the AU detection task, which formula is:
\begin{align}
\hat{y}^{AU} = \frac{1}{1+e^{-y_{pre}^{AU}}}
\end{align}
\begin{align}
\mathcal{L}^{AU} = \frac{1}{N}\sum\limits_{i}-(y_i^{AU}\log(\hat{y}_i^{AU})+(1-y_i^{AU})\log(1-\hat{y}_i^{AU}))
\end{align}
where $y_{pre}^{AU}$ denotes the AU prediction result of the model, $y_i^{AU}$ denotes the true label for AU of class $i$.

For the EXPR task, we use the cross-entropy loss as the inference loss, with the following equation:
\begin{align}
\hat{y}_{i}^{EXPR} = \frac{e^{y_{pre(i)}^{EXPR}}}{\sum\limits_{k=1}\limits^{K}e^{y_{pre(k)}^{EXPR}}}
\end{align}
\begin{align}
\mathcal{L}^{EXPR} = -\sum\limits^{K}\limits_{i=1}y_i^{EXPR}\ln{\hat{y}_{i}^{EXPR}}
\end{align}
where $y_{pre(i)}^{EXPR}$ denotes the EXPR prediction result of the model of class $i$, $y_i^{EXPR}$ denotes the true label for EXPR of class $i$.

Finally, we use the negative Concordance Correlation Coefficient (CCC) as the inference loss for the VA prediction, with the following equation:
\begin{align}
\mathcal{L}^{VA} = 1-CCC^{V} + 1-CCC^{A}
\end{align}
The sum of the three loss values is used as the overall evaluation inference loss of the multitask model:
\begin{align}
\mathcal{L} = \mathcal{L}^{AU} + \mathcal{L}^{EXPR} + \mathcal{L}^{VA}
\end{align}

\section{Experiments}
\subsection{Datasets}
Static dataset s-Aff-Wild2 \cite{DBLP:journals/corr/abs-2202-10659,DBLP:conf/eccv/KolliasCPZ18,DBLP:journals/ijcv/KolliasCVKZ20,DBLP:conf/cvpr/KolliasNKZZ17,DBLP:journals/corr/abs-2105-03790,DBLP:journals/ijcv/KolliasTNPZSKZ19,DBLP:conf/bmvc/KolliasZ19,DBLP:conf/acivs/KolliasZ20,DBLP:journals/corr/abs-2103-15792,DBLP:conf/cvpr/ZafeiriouKNPZK17},  for multi-task learning (MTL) \cite{kollias2022abaw}, is provided by the ABAW4 competition. It contains only a subset of frames from the Aff-Wild2 dataset. Each frame has a corresponding AU, EXPR and VA labels. 

However, there are abnormal labels in this dataset, the cases of $-1$ for EXPR label, $-5.0$ for VA label, and $-1$ for both AU labels, respectively. In our experiments, to balance the number of samples between multiple tasks, we did not use the data with abnormal labels in the training and validation phase.

\subsection{Training Details}
The aligned faces are provided by this Challenge. Each image has a resolution of $112\times112$. We resize it to $224\times224$ and feed it into the model.
In the ROI Feature Extraction module, we try different settings and choose an implementation with U of 24 and D of 24. Transformer blocks share the same architecture, containing 4 attention head. The hidden units of the feed-forward neural network in these blocks are set to be 1024.
The optimizer we used is Adam, and the total number of training epochs is 100. 

\subsection{Evaluation Metric}

We use the averaged F1 score of 12 AUs as the evaluation score for AU detection, the averaged macro F1 score as the evaluation score for EXPR prediction, and the CCC value as the evaluation score for the VA task.

\section{Results}
\subsection{Comparison With Baseline Models}
In the ABAW4 challenge, the official baseline model is provided which use pre-trained VGG16 network to extract features, and 22 simple classifiers for multitask(2 linear units for VA predictions, 8 softmax activation function units for EXPR predictions and 12 sigmoid activation function units for AU predictions). 
The experiment results are shown in table \ref{results}, demonstrate the superiority of our model over the baseline model.

\subsection{Different Temporal Smoothing Methods}
We try to add a Bidirectional Temporal Smoothing operation (\emph{BTS}) to each task, but only assign a trainable initial state vector to the forward operation, i.e. the reverse operation all starts from the penultimate time step.
The experiment results are shown in table \ref{results}, it is not as effective as Temporal Smoothing operation (\emph{TS}).

\subsection{Different Backbones}
We try to use the same structure as Sign Vehicle Space and Message Space of the SMM-EmotionNet \cite{DBLP:journals/corr/abs-2203-12845} (\emph{SMM}) to extract feature vectors separately before feeding them into the Temporal Smoothing and Classifier Module, but the results are not satisfactory.

We try to use only one Transformer block (\emph{Single Perspective}) in the Interaction Module to extract information, and the performance degradation is significant compares to two Transformer blocks (\emph{Double Perspective}). This also proves that the facial emotion descriptors information extracted by two Transformer blocks is more complete.
The experiment results are shown in table \ref{results}.

\begin{table}[]
	\centering
	\begin{tabular}{ccccccc}
		\hline
		Model                                      & TS & BTS & SMM & \begin{tabular}[c]{@{}c@{}}Single\\ Perspective\end{tabular} & \begin{tabular}[c]{@{}c@{}}Double\\ Perspective\end{tabular} & Overall Metric \\ \hline
		\multicolumn{1}{c|}{baseline}              & -  & -   & -          & -    & -    & 0.3            \\ \hline
		\multicolumn{1}{c|}{\multirow{4}{*}{ours(U=17,D=16)}} &    & \checkmark   &            &      & \checkmark    & 0.759          \\ \cline{2-7} 
		\multicolumn{1}{c|}{}                      & \checkmark  &     & \checkmark          &      &      & 0.815           \\ \cline{2-7} 
		\multicolumn{1}{c|}{}                      & \checkmark  &     &            & \checkmark    &      & 0.79          \\ \cline{2-7} 
		\multicolumn{1}{c|}{}                      & \checkmark  &     &            &      & \checkmark    & 0.825           \\ \hline
		\multicolumn{1}{c|}{ours(U=24,D=24)}                  & \checkmark  &     &            &      & \checkmark    & \textbf{0.85}          \\ \hline
	\end{tabular}
	\caption{Results of all experiments on the validation set.}
	\label{results}
\end{table}

\subsection{Different semantic information}
\begin{figure}
	\centering
	\includegraphics[width=0.5\linewidth]{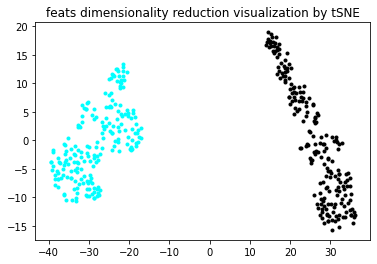}
	\caption{Outputs of two Transformer blocks, the sample points of different colors represent the outputs of different Transformer blocks.}
	\label{fig:outputs}
\end{figure}

To verify the effectiveness of these two Transformer blocks, we extract some samples on the validation set for inference and use the t-SNE algorithm for dimensionality reduction to visualize the outputs of these two blocks. The results are shown in Figure \ref{fig:outputs}.

The results show that the sample points of the two colors are clearly demarcated, indicating that there are different types of perspective information extracted from the two Transformer blocks.
\subsection{Additional}
The performance of our model on each task is shown in Table \ref{tasks}.

\begin{table}[]
	\centering
	\begin{tabular}{c|cccc}
		\hline
		Task            & Valence & Arousal & Expression & Action Units \\ \hline
		ours(U=24,D=24) & 0.41    & 0.62    & 0.207      & 0.385        \\ \hline
	\end{tabular}
	\caption{Results of all tasks.}
	\label{tasks}
\end{table}

\section{Conclusions}
In this paper, we introduce our method for the Multi-Task Learning Challenge ABAW4 competition. We extracted different perspective information for Sign Vehicle space and Message space, and enhanced the model's utilization of these information by enabling multi-task information to interact through an attention mechanism. The results show that our method achieves a performance of 0.85 on the validation set.

\clearpage
% ---- Bibliography ----
%
% BibTeX users should specify bibliography style 'splncs04'.
% References will then be sorted and formatted in the correct style.
%
\bibliographystyle{splncs04}
\bibliography{egbib}
\end{document}